\documentclass[runningheads]{llncs}
\usepackage{graphicx}
\usepackage{amsmath}
\usepackage{amssymb}
\usepackage{booktabs}
\usepackage{bbding}
\usepackage{wrapfig}
\begin{document}
\title{Robust ID-Specific Face Restoration via Alignment Learning}
%
%

\author{Yushun Fang\inst{1}${\dagger}$ \and
Lu Liu\inst{1}${\dagger}$ \and 
Xiang Gao\inst{1}${\dagger}$ \and 
Qiang Hu\inst{1(}\Envelope\inst{)} \and 
Ning Cao\inst{2} \and 
Jianghe Cui\inst{2} \and 
Gang Chen\inst{2} \and 
Xiaoyun Zhang\inst{1(}\Envelope\inst{)}}

\authorrunning{Fang et al.}

\institute{Shanghai Jiao Tong University, Shanghai, 200240, China \and
E-surfing Vision Technology Co., Ltd., China}

\maketitle 
\footnotetext[1]{Equal contribution: Yushun Fang, Lu Liu, and Xiang Gao equally contributed to this study.}
\footnotetext[2]{Corresponding authors: Qiang Hu (e-mail: qiang.hu@sjtu.edu.cn) and Xiaoyun Zhang (e-mail: xiaoyun.zhang@sjtu.edu.cn).}

\begin{abstract}
The latest developments in Face Restoration have yielded significant advancements in visual quality through the utilization of diverse diffusion priors. Nevertheless, the uncertainty of face identity introduced by identity-obscure inputs and stochastic generative processes remains unresolved. To address this challenge, we present Robust ID-Specific Face Restoration (RIDFR), a novel ID-specific face restoration framework based on diffusion models. Specifically, RIDFR leverages a pre-trained diffusion model in conjunction with two parallel conditioning modules. The Content Injection Module inputs the severely degraded image, while the Identity Injection Module integrates the specific identity from a given image. Subsequently, RIDFR incorporates Alignment Learning, which aligns the restoration results from multiple references with the same identity in order to suppress the interference of ID-irrelevant face semantics (e.g. pose, expression, make-up, hair style). Experiments demonstrate that our framework outperforms the state-of-the-art methods, reconstructing high-quality ID-specific results with high identity fidelity and demonstrating strong robustness.
\end{abstract}

\begin{figure*}[ht]
      \vspace{-10pt}
  \includegraphics[width=\textwidth]{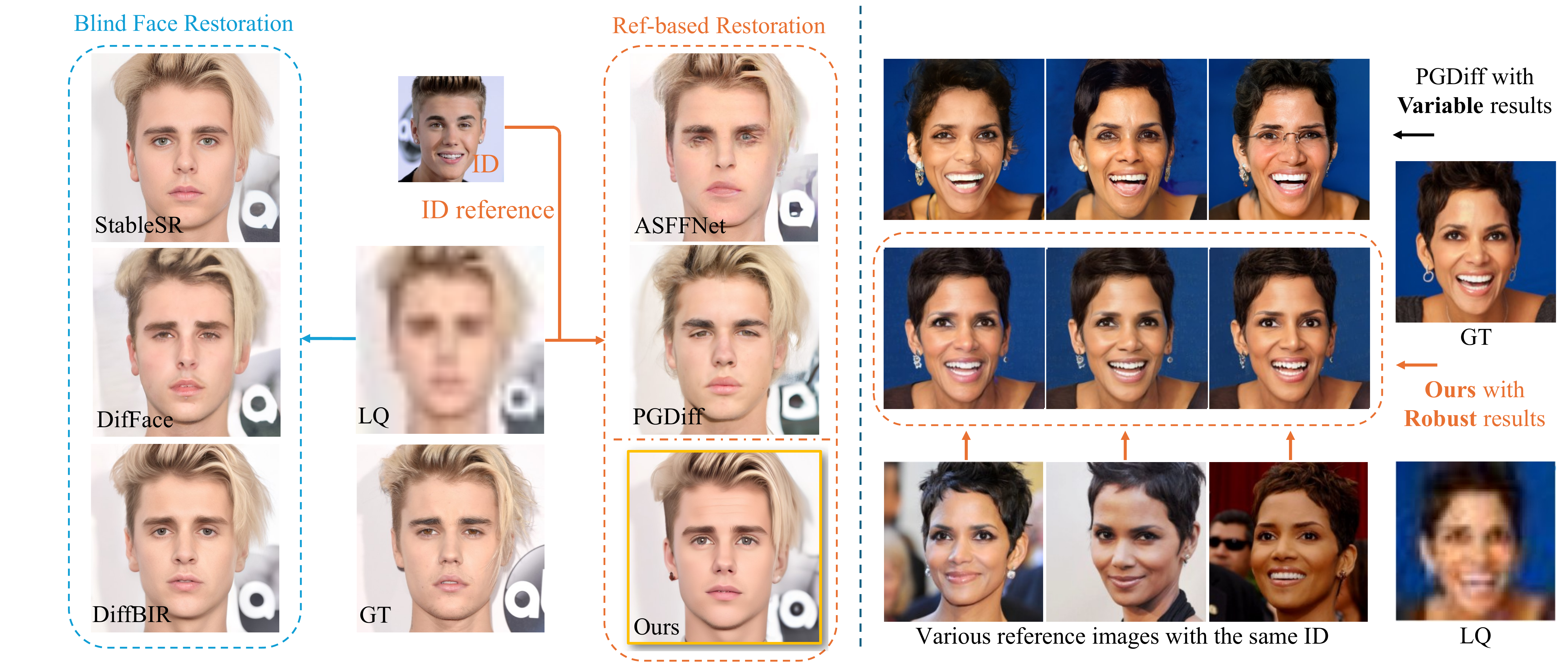}
      \vspace{-23pt}
  \caption{\textit{Left:} For low quality input images, especially those severely degraded and identity obscure, existing blind face restoration methods (StableSR~\cite{wang2023exploiting}, DifFace~\cite{yue2022difface}, DiffBIR~\cite{lin2023diffbir}) often lead to identity unexpected results, while reference-based restoration methods suffer from the robustness issue in inaccurate identity (PGDiff~\cite{yang2024pgdiff}), low image quality or requiring extra conditions such as face landmarks(ASFFNet~\cite{li2020enhanced}). \textit{Right:} Our proposed framework effectively retains both identity fidelity and visual quality with a single-identity reference image, and is robust to any given reference images with various pose, expression and make-up etc.}
    \vspace{-8pt}
  \label{fig:teaser}
\end{figure*}

\section{Introduction}
\label{sec:intro}

Face restoration focuses on reconstructing high-quality (HQ) face images from low-quality (LQ) counterparts, usually originated with complex degradation, such as blur, noise, compression, down-sampling, etc. It is an inherently ill-posed problem, which means that a single LQ face image can correspond to infinite plausible HQ images. Recently, diffusion priors, with the impressive ability to generate abundant and realistic details, have been explored as a prevalent approach in face restoration to compensate for the loss of details in LQ images with visual consistency and semantic coherence{~\cite{yue2022difface,lin2023diffbir,wang2023exploiting}}. 

The existing research leveraging diffusion priors can be categorized into two classes: conditioned finetuning and zero-shot guidance. The former~\cite{lin2023diffbir,wang2023exploiting} insert LQ images as condition to finetune from a large-scale pre-trained diffusion model; The later ~\cite{yue2022difface,yang2024pgdiff} introduce additional constraints from LQ images to guide the de-noising process without re-training. Despite these advancements, the inherently stochastic and versatile generative process often introduces fabricated details, textures, and face attributes. 

This leads to the notable issue of identity fidelity as shown in Figure~\ref{fig:teaser} \textit{Left}, where the restored images of blind face restoration methods~\cite{wang2023exploiting,yue2022difface,lin2023diffbir} fail to consist with an expected identity. To address this issue, reference-based methods~\cite{liu2025faceme,yang2024pgdiff,li2020enhanced,li2022learning} introduce HQ reference images or extra conditions such as face land-marks, but struggling to align high-level face semantics (including identity) from HQ reference images with the LQ input, still often leading to visual artifacts due to the gap between HQ references and LQ input. What's more, various reference images from the same identity may have different pose, expression, make-up or hair style as shown in Figure~\ref{fig:teaser}  \textit{Right}, which further poses great challenge to the restoration robustness. In summary, identity fidelity and quality robustness are still two main issues of the current generative face restoration, especially for severely degraded and identity obscure input images.

In this work, motivated by the observation above and thanks to diffusion priors' ability to generate high-quality details, we propose an ID-specific face restoration framework named \textit{RIDFR}, aiming to inject specific identity from a given ID image into the restoration process, while being robust to ID images' different quality and ID-irrelevant face semantics (e.g. pose, expression, make-up, hair style). We leverage a pre-trained diffusion model collaborating with two parallel conditioning modules. Specifically, pre-trained diffusion model is utilized due to its superior capability to generate abundant details, while two parallel conditioning modules -  \textit{Content Injection Module} and  \textit{Identity Injection Module} - work in tandem to utilize identity-obscure LQ images and specific identity for restoration. Then, \textit{Alignment Learning} is introduced to suppress the interference of ID-irrelevant face semantics, leading to more robust restoration performance. 

To summarize, we make the following contributions:
\begin{itemize}
\item We propose RIDFR, a robust framework for ID-specific face restoration achieving collaboration between a diffusion prior and two parallel conditioning modules.
\item We introduce Alignment Learning to align the restoration results from multiple references with the same identity, suppressing the interference of ID-irrelevant face semantics and achieving strong robustness. 
\item Extensive experiments demonstrate our superiority in achieving both identity fidelity and visual quality when restoring identity-obscure LQ images, outperforming existing blind face restoration methods with over 50\% increment in the identity fidelity.
\end{itemize}

\section{Related Work}
\label{gen_inst}

\noindent \textbf{Blind Face Restoration.}
Blind face restoration aims to restore a HQ face image given a degraded LQ image. Plenty of facial priors are utilized to alleviate dependency on the input LQ image, including geometric priors ~\cite{chen2018fsrnet,kim2019progressive}, Vector-Quantized Codebook prior~\cite{zhou2022towards,gu2022vqfr}, etc. Currently, generative priors, such as GAN priors~\cite{yang2021gan,wang2021gfpgan} and diffusion priors~\cite{yue2022difface,lin2023diffbir,wang2023dr2,wang2023exploiting}, are utilized to provide vivid details and texture. Diffusion priors have been widely adopted in recent research including DiffBIR~\cite{lin2023diffbir}, StableSR~\cite{wang2023exploiting},DifFace~\cite{yue2022difface}, which enrich the restoration with abundant details while poses significant challenges to fidelity issues like identity uncertainty. Ours framework introduces specific identity to address this issue, improving practicality of face restoration methods.

\noindent \textbf{Reference-Based Face Restoration.}
Besides the aforementioned priors, HQ reference images can compensate for LQ inputs by transferring texture ~\cite{lau2024ented}, details~\cite{Li_2018_ECCV,Dogan_2019_CVPR_Workshops}, and identity ~\cite{li2020enhanced,li2022learning}. These methods can be classified into three categories: (i) multi-reference methods. Previous work including ASFFNet~\cite{li2020enhanced}, DMDNet~\cite{li2022learning} employ multi-reference images as input. These methods strongly rely on face landmarks extracted from LQ images for accurate alignment between reference images and LQ images, which limits their application in severe degraded situations. Utilizing diffusion priors, PFStorer~\cite{varanka2024pfstorer} is a few-shot methods requiring test-time optimization with multiple images for each given identity, resulting in low computation efficiency. FaceMe~\cite{liu2025faceme} support both multi- and single-reference inputs, however, it relies on a large amount of additional synthesized data for training and a much larger base model as prior, which poses challenges in terms of both image quality and computation required.
(ii) single-reference methods. PGDiff~\cite{yang2024pgdiff} leverages a pre-trained diffusion model by formulating zero-shot guidance for universal image restoration. For blind face restoration, it adapts the classifier guidance using ArcFace~\cite{deng2019arcface} loss across various degrees of image degradation, leading to unstable restoration with inappropriate edges and misaligned facial attributes. MGFR~\cite{tao2024overcoming} utilizes high-quality reference image and prompt for better restoration quality, but its reliance on additional models like facial attribute extractor compromises computation efficiency and requires much more high-quality training data. Different from these methods, our framework focuses on a single ID-recognizable face image as reference with no test-time optimization, while utilizing basic diffusion prior and limited amount of training data.

\begin{figure*}[ht]
\centering
\vspace{-20pt}
\includegraphics[width=1.0\linewidth, trim=0 0 0 0]{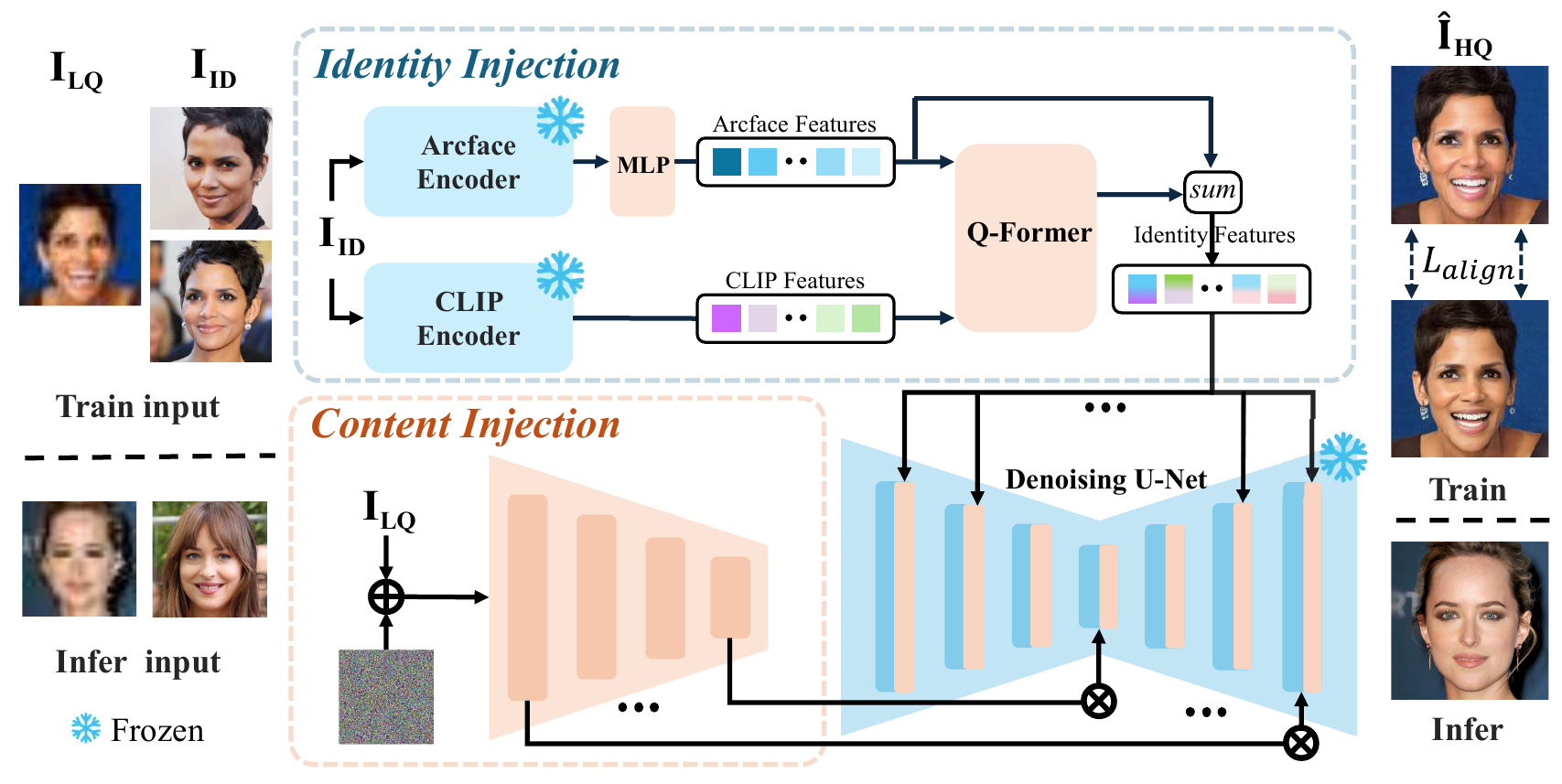}
      \vspace{-23pt}
\caption{\textbf{Overall framework}.The low-quality image $I_\text{LQ}$ and identity image  $I_\text{ID}$  are injected by the \textit{Content Injection Module} and \textit{Identity Injection Module} in parallel to guide the pre-trained diffusion model. Alignment Learning is introduced to align the restoration results (the predicted noises, to be precise) from multiple references with the same identity to suppress the ID-irrelevant interference. For inference, only a single identity image is needed for robust ID-specific restoration.}
\label{Fig:framwork}
\vspace{-15pt}
\end{figure*}

\noindent \textbf{Identity Customization for Image Generation.}
Recently, plenty of research concerned about identity customization methods for image generation~\cite{ye2023ip,li2023photomaker,cui2024idadapter,guo2024pulid}, which aims to generate creative images of individuals with the input of their single or multiple face images.  This is closely related to identity-specific face restoration, sharing the same key idea of how to inject identity information during the generation process.  One common way is to introduce conditioning modules based on diffusion models, and finetune the image encoders or identity adapters with an ID-oriented dataset. PhotoMaker~\cite{li2023photomaker}, and InstantID~\cite{wang2024instantid} extract the specific ID embedding by finetuning the image encoder. IP-Adapter-Face~\cite{ye2023ip} and IDAdapter~\cite{cui2024idadapter} use a frozen face recognition model and CLIP image encoder to retain the identity through finetuning the adapters. But as PuLID~\cite{guo2024pulid} points out, both two ways introduce ID-irrelevant face semantics (e.g. pose, expression, make-up, hair style), thus weakening models' diversity in image generation. In this paper, inspired by previous works, we enhance the identity injection performance with proposed \textit{Alignment Learning} to align the restoration results from multiple references with the same identity in order to suppress the interference of ID-irrelevant face semantics.

\section{Method}

\label{headings}

\subsection{Overview}
Our proposed RIDFR framework is illustrated in Figure \ref{Fig:framwork}. Given an ID-obscure LQ face image $I_\text{LQ}$, our framework aims to restore an ID-specific face $\hat{I}_\text{HQ}$ by embedding ID features from any given identity image $I_\text{ID}$, formulated as follows:
\begin{equation}
    \hat{I}_\text{HQ} = \Omega(\theta, I_\text{LQ}, I_\text{ID})
    \label{eg:framework}
\end{equation}
Here, $\theta$ denotes the pre-trained diffusion prior. The restored face $\hat{I}_\text{HQ}$ should have high visual quality and the specific identity same with the given $I_\text{ID}$. It is noteworthy that $I_\text{ID}$ only needs to have recognizable identity, and its visual quality, which may required by other reference-based restoration methods for texture or other detail transfer, is not essentially required in our method.

To incorporate $I_\text{LQ}$ and $I_\text{ID}$ as conditions into pre-trained diffusion, our framework introduces Content Injection Module and Identity Injection Module respectively. Content Injection Module extracts low-level image content features from $I_\text{LQ}$ and guides the diffusion prior to generate detail enriched face image, but suffering uncertain or unexpected identity. Identity Injection Module thus embeds high-level identity features from $I_\text{ID}$ and ensure the specific identity or ID fidelity for restoration. To suppress the interference from ID-irrelevant semantics of $I_\text{ID}$ and mitigate conflicts between the two modules, Alignment Learning is introduced to align the restoration results from multiple references with the same identity. In inference, for an ID-obscure LQ input, by providing a single identity image as reference, our method is capable of ID-specific face restoration and robust to variants of the reference image such as different pose, expressions, make-up and hair style refer to experiments (Section \ref{exp:compare}) .

\begin{figure*}[ht]
\centering
\includegraphics[width=1.0\linewidth, trim=0 0 0 0]{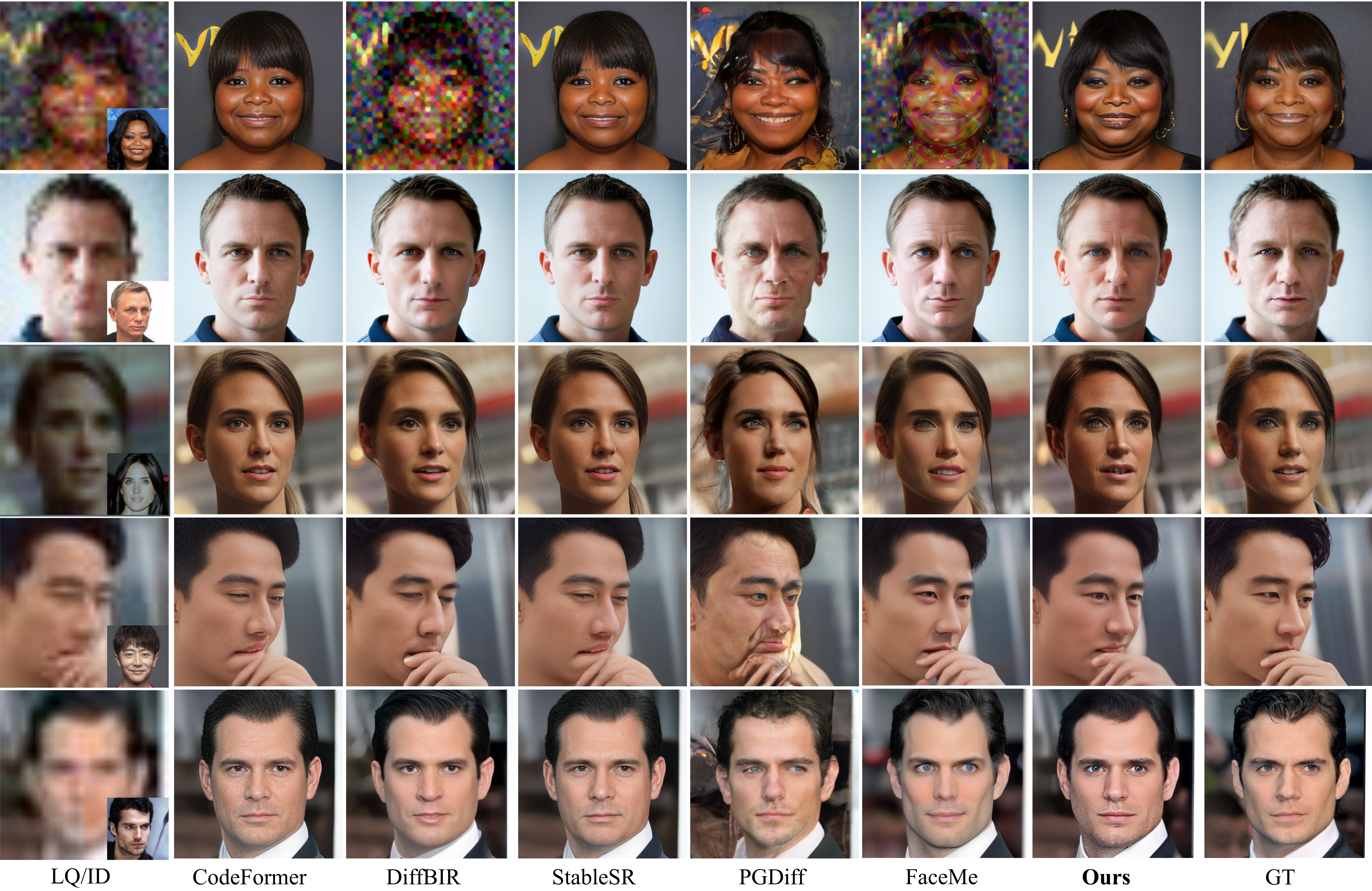}
\vspace{-20pt}
\caption{Qualitative Comparison on \textit{CelebRef-HQ-Test}. Our method achieves superior identity fidelity and visual quality.}
\label{Fig:x16}
\end{figure*}

\subsection{Content and Identity Injection}
\noindent \textbf{Content Injection Module.}
\label{cim}
is designed to address the basic face restoration task, producing $\hat{I}_\text{HQ}$ merely based on $I_\text{LQ}$ without specific ID. Here, we follow ControlNet~\cite{zhang2023adding} with key changes of condition encoding: we concatenate $I_\text{LQ}$ in pixel space with the input noisy latent $z_t$ along the channel axis and increase the channel of input convolution layers from 4 to 7. Two main considerations are taken into account: (i) compared with adding the encoded condition to the input noisy latent in ControlNet, the modification in input channel can avoid the severe color shift problem ~\cite{lin2023diffbir}. (ii) pixel-level concatenation, compared with latent-level concatenation adopted in ~\cite{lin2023diffbir} can avoid the information loss caused by encoding $I_\text{LQ}$ into latent space, which is demonstrated in the ablation study \ref{abl:concat}. This module is trained in the same way as training ControlNet with Diffusion models~\cite{zhang2023adding} by minimizing:
\begin{equation}
    \mathcal{L}_{\text{Content}}= \mathbb{E}_{z_t, t, \epsilon}[||\epsilon-\epsilon_{\theta}(z_t, t, I_\text{LQ})||^2_2],
\end{equation}

\noindent \textbf{Identity Injection Module.}
\label{iim}
aims to extract and inject the identity information of $I_\text{ID}$, composing an ArcFace~\cite{deng2019arcface} encoder, a CLIP~\cite{radford2021learning} encoder and a Q-Former~\cite{li2023blip}. Specifically, Arcface feature is projected using an MLP to serve as \textit{query} for Q-Former to extract related CLIP feature, and a shortcut structure is employed to produce the final identity feature $c_\text{ID}$. The feature is subsequently injected into the U-Net model following decoupled cross-attention mechanism proposed in ~\cite{ye2023ip}:
\begin{equation}
    \mathrm{Attention}(Q, K, V)+ \mathrm{Attention}(Q, K', V') \label{eq:ip-adapter}
\end{equation}
Here, $Q = \phi(z_t)W_Q$ is query from the intermediate feature $\phi(z_t)$ of the U-Net model. $K = c_\text{text}W_K$ and $V = c_\text{text}W_V$ are key and value from text embedding $c_\text{text}$. $K' = c_\text{ID}W_K'$ and $V' = c_\text{ID}W_V'$ are key and value from the additional identity feature $c_\text{ID}$. $W_K$ and $W_V$ are frozen projection matrices from the U-Net model while $W_K'$ and $W_V'$ are learn-able and newly injected. It is worth noting that Identity Injection Module is not limited to any particular design of identity encoder, which will be further discussed in Section ~\ref{method:align}.

\subsection{Alignment Learning}
\label{method:align}
While the Content Injection Module makes full use of the content in LQ image and the Identity Injection Module extracts sufficient identity, their direct collaboration can roughly maintain both identity fidelity and visual quality (Section \ref{abl:align}). But as shown in Figure~\ref{fig:teaser}, for ID-specific face restoration task, ideally we want to get restored results consistent with a certain LQ input across various ID images as long as these images share the same identity. However, no face identity encoder is capable of extracting exact the same ID embedding from these images, meaning that ID-irrelevant semantics (e.g. pose, expression, make-up, hair style) always exist. Besides, training identity encoders based on diffusion models often incorporates self-reconstruction of input images, which will make full use of face semantics and result in more serious interference from ID-irrelevant semantics~\cite{guo2024pulid}. In ID-specific face restoration task, this interference leads to uncertainty and conflicts with content injection from LQ images. For instance, if the LQ image depicts a closed mouth, but the ID image shows a laughing face, the restoration should focus on accurately reconstructing the closed mouth based on the LQ image identity-specifically, rather than merely replicating the laugh from the ID image. 

\begin{figure*}[ht]
    \centering
    \vspace{-10pt}
    \includegraphics[width=1\linewidth]{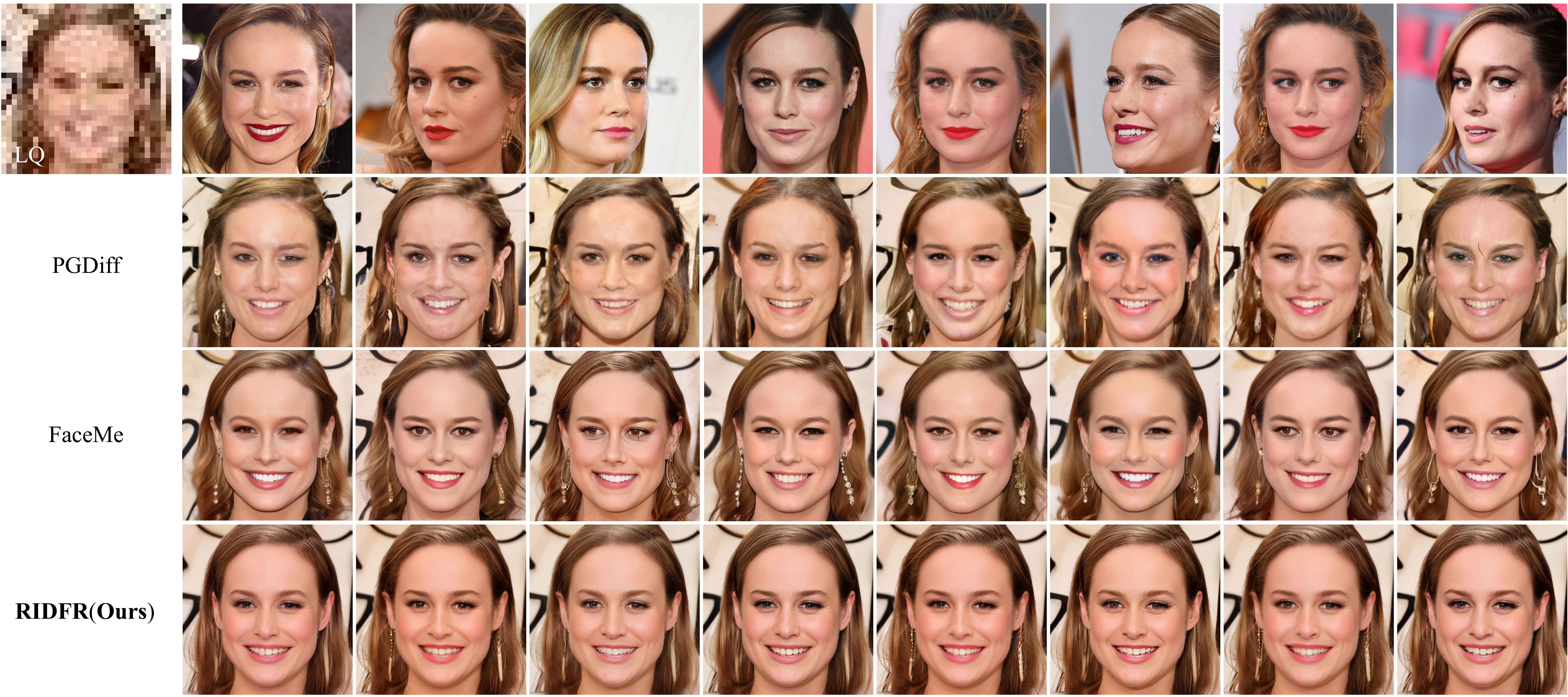}
    \vspace{-20pt}
    \caption{Robustness to various reference images. Our proposed method can restore consistent results when the provided ID images have different poses, expressions, make-up, hair style and illumination.}
    \label{fig:abl_ridfr_pgdiff}
    \vspace{-15pt}
\end{figure*}

To mitigate the ID-irrelevant semantics from ID images and enhance consistency with LQ input, we employ a finetuning-after-training strategy, \textit{i.e.,} Alignment Learning, to finetune the framework after training Content Injection Module, aligning the predicted noises of the U-Net model under $<I_\text{LQ},I_\text{HQ},I_{\text{ID}_{\{1,2\}}}>$ pairs where $I_{\text{ID}_{\{1,2\}}}$ are randomly chosen from all reference images with the same identity. In this way, the framework learns local (two random images) and global (all reference images) alignment asynchronously under the same identity, which achieves a balance between performance and tuning cost. The objective of Alignment Learning can be expressed as follow:
\begin{equation}
    \mathcal{L}_{\text{align}} = \mathbb{E}_{z_t, t, \epsilon}[||\hat{\epsilon_{\text{ID}_1}}-\hat{\epsilon_{\text{ID}_2}}||^2_2],
\end{equation}
Here, $\hat{\epsilon_{\text{ID}_1}}$ and $\hat{\epsilon_{\text{ID}_2}}$ represent predicted noise given ID images $I_{\text{ID}_1}$ and $I_{\text{ID}_2}$. We also incorporate original Diffusion training objective as follows:
\begin{equation}
    \mathcal{L}_{\text{Diffusion}} = \mathbb{E}_{z_t, t, \epsilon}[||\epsilon-\epsilon_{\theta}(z_t, t, I_\text{LQ}, I_{\text{ID}_{\{1,2\}}})||^2_2],
\end{equation}
The overall objective of Alignment Learning can be expressed as:
\begin{equation}
    \mathcal{L} = \mathcal{L}_{\text{Diffusion}} + \lambda\mathcal{L}_{\text{align}}
\end{equation}
Where $\lambda$ is the hyper-parameter controlling the importance of alignment learning. We observe that high $\lambda$ value harms identity fidelity, guiding the framework to align results ignoring the injected identity. So we set $\lambda$ to 1.0 in practice. Besides, to further reduce tuning cost, we initialize the weight of Identity Injection Module using an IP-Adapter-FaceID-Plus checkpoint\footnote{https://huggingface.co/h94/IP-Adapter-FaceID} and keep it frozen to reduce learnable parameters, which leads to no performance drop of the framework according to our observations, while ensuring flexibility that Identity Injection Module is not limited to any particular design of identity encoder.

\section{Experiments}

\label{experiment}

\subsection{Experiment Settings}
\noindent \textbf{Implementation.}
We choose Stable Diffusion v1.5 as the pretrained diffusion prior of our framework. We train Content Injection Module for 87,500 iterations with batch size of 8, and finetune Content Injection Module with Alignment Learning for 45,000 iterations with batch size of 8. For inference, UniPC~\cite{zhao2024unipc} sampling with 20 steps is adopted and a pre-trained SwinIR~\cite{yue2022difface,lin2023diffbir} is employed to pre-process $I_\text{LQ}$ for fair comparison. Notably, our framework do not rely on any specific pre-processing. Besides, in Content Injection Module, bicubic up-sampling to $64 \times 64$ is adopted to $I_\text{LQ}$ initially to match the size of the U-Net model's input latent. Prompt condition is set to empty both in training and finetuning stage.

\noindent \textbf{Datasets.}
We train Content Injection Module on FFHQ~\cite{karras2019style}, containing 70,000 high-quality images with $512 \times 512$ resolution, and finetune Content Injection Module on CelebRef-HQ~\cite{li2022learning}, containing 1,005 identities and 10,555 high-quality images with $512 \times 512$ resolution in total, $\{3, \dots,21\}$ images for each identity. We randomly select 905 identities with 9,453 images as the finetuning dataset. 

During training and fine-tuing, $<I_\text{LQ},I_\text{HQ},I_{\text{ID}_{\{1,2\}}}>$ pairs are constructed online with synthesized LQ images following common degradation model expressed in Eq.\ref{Eq:deg} following ~\cite{wang2021real} and randomly down-sampled ID images for simulating ID-recognizable images with uneven quality. 
\begin{equation}
    \vspace{-0.5mm}
    I_\text{LQ} = [(I_\text{HQ} \circledast k_\sigma) \downarrow_r + n_\delta]_{JPEG_q}
    \label{Eq:deg}
\end{equation}

For identity images $I_\text{ID}$, we randomly down-sample original high-quality reference images to resolution between $512 \times 512$ and $64 \times 64$. 

For testing, we construct a synthetic dataset and a real-world dataset. First, we use the remaining 100 identities and 1,102 images from CelebRef-HQ as \textit{CelebRef-HQ-Test}. For real-world cases, we extract a small subset of LFW ~\cite{huang2008labeled,huang2012learning}, denoted as \textit{LFW-Ref-Test}, using corresponding reference images from CelebRef-HQ-Test. LFW-Ref-Test contains 16 identities and 76 images in total. Other details are presented in supplementary materials.

\begin{table*}[t]
    \vspace{-5pt}
  \caption{Quantitative comparisons on \textit{CelebRef-HQ-Test}. The best and the second-best are highlighted in \textbf{bold} and \underline{underline}. The upper rows are non-diffusion-prior-based methods, the mid rows are diffusion-prior-based methods and the lower rows are diffusion-prior-based methods with reference.}
    \vspace{-5pt}
  \label{Tab:x16}
  \centering
  \resizebox{0.88\textwidth}
  {!}{
      \begin{tabular}{c|c c|c c c c c}
        \toprule
         Methods & LPIPS$\downarrow$ & IDS-HQ$\uparrow$ & MANIQA$\uparrow$ & MUSIQ$\uparrow$ & CLIP-IQA$\uparrow$ & HyperIQA$\uparrow$ & PI$\downarrow$  \\
        \midrule
        GFPGAN\cite{wang2021gfpgan} & 0.286 & 0.360 & 0.599 & 75.363 & 0.695 & 0.744 & 3.564 \\ 
        GPEN\cite{yang2021gan} & 0.307 & \underline{0.389} & 0.568 & 75.423 & 0.666 & 0.728 & \underline{3.072} \\ 
        CodeFormer\cite{zhou2022towards} & \textbf{0.269} & 0.376 & 0.577 & 76.102 & 0.708 & 0.744 & 3.170 \\ 
        VQFR\cite{gu2022vqfr} & 0.281 & 0.368 & 0.564 & 73.212 & 0.626 & 0.707 & 4.753 \\ 
        Face-SPARNet\cite{chen2020learning} & 0.291 & 0.363 & 0.515 & 71.740 & 0.620 & 0.677 & 3.870 \\ 
        \midrule
        DiffBIR\cite{lin2023diffbir} & 0.305 & 0.373 & \underline{0.634} & 73.806 & \underline{0.723} & 0.750 & 4.203 \\ 
        DifFace\cite{yue2022difface} & 0.306 & 0.314 & 0.481 & 66.651 & 0.553 & 0.660 & 5.416 \\ 
        StableSR\cite{wang2023exploiting} & \underline{0.270} & 0.357 & 0.607 & \underline{76.138} & 0.698 & \underline{0.759} & 3.575 \\ 
        DR2\cite{wang2023dr2} & 0.319 & 0.289 & 0.548 & 70.446 & 0.626 & 0.707 & 4.753 \\
        \midrule
        PGDiff\cite{yang2024pgdiff} & 0.325 & \textbf{0.549} & 0.454 & 66.934 & 0.555 & 0.651 & 4.438 \\ 
        FaceMe\cite{liu2025faceme} & 0.280 & 0.500 & 0.553 & 74.349 & 0.660 & 0.712 & 4.121 \\ 
        \midrule
        RIDFR(Ours) & 0.294 & \textbf{0.549} & \textbf{0.656} & \textbf{76.826} & \textbf{0.782} & \textbf{0.776} & \textbf{2.999} \\ 
       
        \bottomrule
      \end{tabular}
  }
  \vspace{-7pt}
\end{table*} 

\noindent \textbf{Metrics.}
For evaluation, we utilize two categories of metrics to evaluate image quality and identity fidelity respectively. (1) The perceptual metrics include LPIPS~\cite{zhang2018lpips}, MANIQA~\cite{yang2022maniqa}, MUSIQ~\cite{ke2021musiq}, CLIP-IQA~\cite{wang2022exploring}, HyperIQA~\cite{Su_2020_CVPR} and PI~\cite{blau20182018}.
(2) The identity metric is measured by Identity Similarity Score (IDS). IDS measures the cosine similarity of ArcFace identity features between the paired images $<\hat{I}_\text{HQ},I_\text{HQ}>$ (IDS-HQ). Here, we adopt an ArcFace model with an alternative backbone and training dataset different from what we employ in Identity Injection Module for fair comparison. 

\begin{figure*}[ht]
    \centering
    \includegraphics[width=0.6\linewidth]{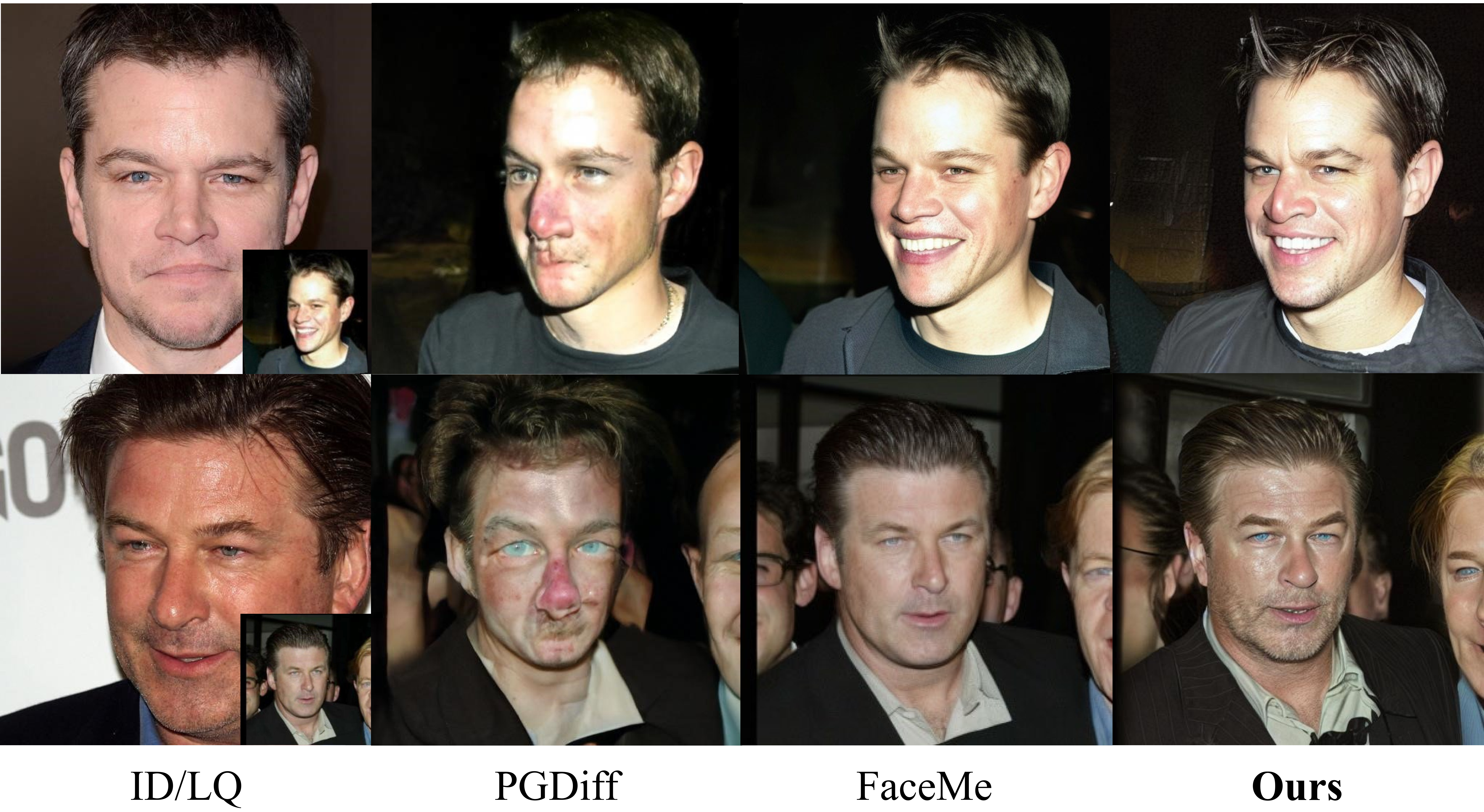}
    \vspace{-10pt}
    \caption{Qualitative results on real world \textit{LFW-Ref-Test}.}
    \label{Fig:lfw}
     \vspace{-15pt}
\end{figure*}

\subsection{Comparisons with State-of-the-art Methods}
\label{exp:compare}
We compare our method with several state-of-the-art methods, including CNN-/Transformer-/GAN-based methods GFPGAN~\cite{wang2021gfpgan}, GPEN~\cite{yang2021gan}, CodeFormer~\cite{zhou2022towards}, Face-SPARNet~\cite{chen2020learning} and diffusion-prior-based methods DiffBIR~\cite{lin2023diffbir}, DifFace~\cite{yue2022difface}, StableSR~\cite{wang2023exploiting}, DR2~\cite{wang2023dr2}, PGDiff~\cite{yang2024pgdiff}, FaceMe~\cite{liu2025faceme}. The last two are also reference-based methods. PFStorer~\cite{varanka2024pfstorer} and MGFR~\cite{tao2024overcoming} are excluded from comparison for no open-sourced codes and models available.

\begin{figure*}[ht]
\vspace{-10pt}
    \centering
    \includegraphics[width=0.9\linewidth]{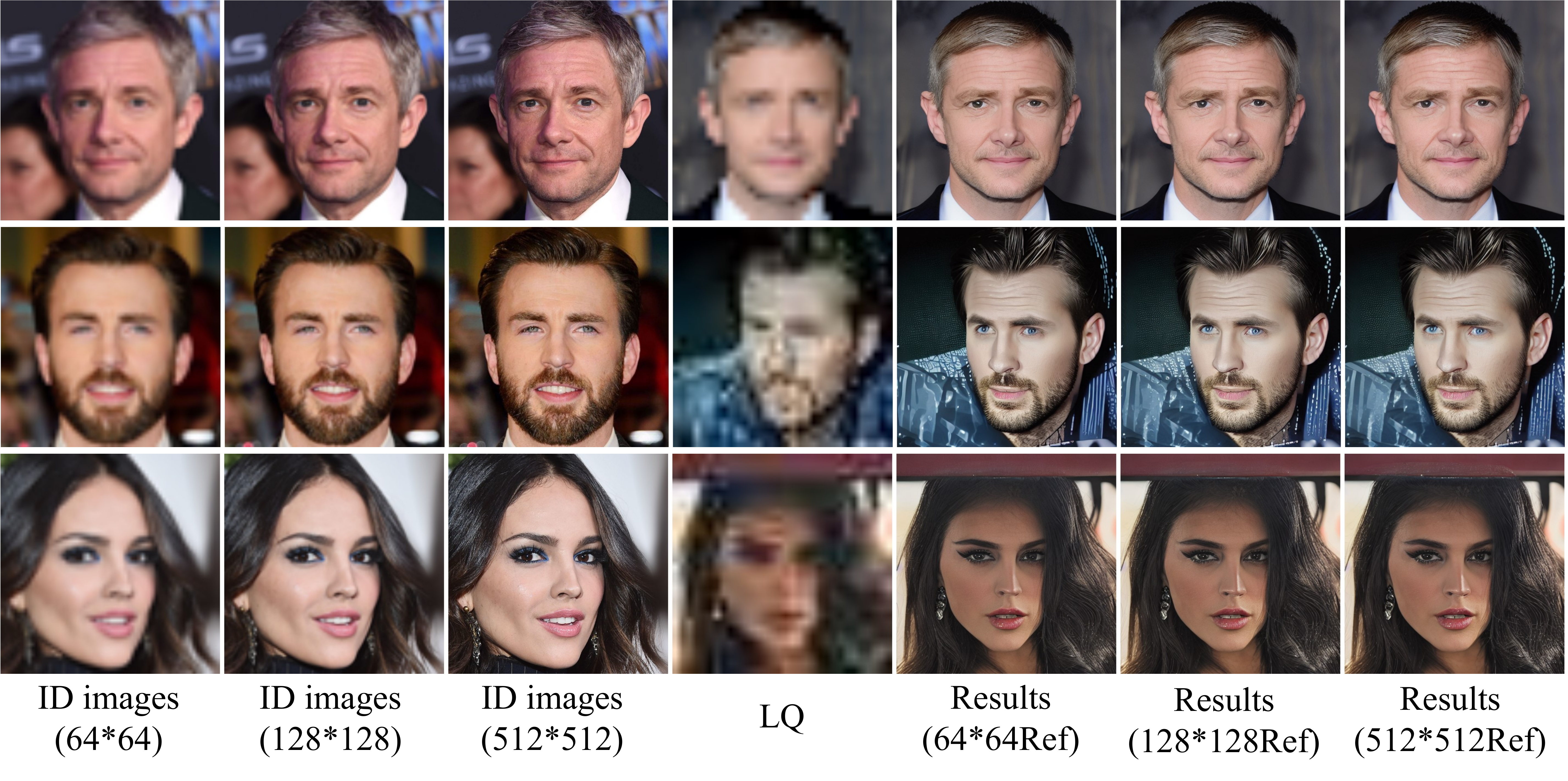}
    \vspace{-15pt}
    \caption{Given ID images with different quality, RIDFR restores consistent high quality images, demonstrating its robustness.}
    \label{fig:different_resolution}
\end{figure*}

\begin{table*}[t]
  \caption{Quantitative results on \textit{CelebRef-HQ-Test} with different resolution ID images. The best and the second-best are highlighted in \textbf{bold} and \underline{underline}. RIDFR can achieve robust results with ID images of different quality.}
  \label{Tab:diff_quality}
  \centering
  \resizebox{0.80\textwidth}
  {!}{
      \begin{tabular}{c|c c|c c c c c}
        \toprule
         Resolution & LPIPS$\downarrow$ & IDS-HQ$\uparrow$ & MANIQA$\uparrow$ & MUSIQ$\uparrow$ & CLIP-IQA$\uparrow$ & HyperIQA$\uparrow$ & PI$\downarrow$  \\
        \midrule
        $64 \times 64$  & 0.301 & 0.531 & 0.651 & 76.649 & 0.783 & \underline{0.766} & 3.079 \\ 
        $128 \times 128$  & \underline{0.300} & \underline{0.548} & \textbf{0.664} & \textbf{76.900} & \textbf{0.794} & \textbf{0.776} & \underline{3.012} \\ 
        \midrule
        Mixed (above) & \textbf{0.294} & \textbf{0.549} & \underline{0.656} & \underline{76.826} & \underline{0.782} & \textbf{0.776} & \textbf{2.999} \\ 
        \bottomrule
      \end{tabular}
  }
  \vspace{-10pt}
\end{table*} 

\noindent \textbf{Comparison on Synthetic Dataset.} Quantitative results are presented in Table~\ref{Tab:x16}. Compared with other state-of-the-art methods, our framework achieves comparable performance in perceptual similarity, qualifying the capability of addressing face restoration task. In terms of image quality, our framework yields the state-of-the-art performance, successfully exploiting the high-quality generation ability of diffusion-prior. On identity-level metrics, our framework shows a significant advantage over non-reference blind face restoration methods, by evaluating restored images with ground truth images. Particularly, compared with PGDiff, which directly uses Arcface identity cosine similarity loss as diffusion sampling guidance, our framework still has the same performance on identity-level metrics while being substantially ahead on perceptual-level metrics; compared with latest referenced-based method FaceMe, our framework performs better in both identity-level metric and image quality. In a nutshell, the proposed RIDFR achieves remarkable performance, ensuring both image quality and identity fidelity on face restoration task.

As shown in Figure~\ref{Fig:x16}, existing blind face restoration methods, including CodeFormer, DifFace, DiffBIR and StableSR, often result in significant identity uncertainty. On the other hand, the reference-based face restoration method PGDiff struggles to restore high-quality images and generates some visual artifacts due to the ID-irrelevant interference from the reference image. Compared with FaceMe, our method produces images more similar to GT while maintaining higher image quality. In comparison, face images generated by our method achieves better visual quality while maintaining the identity accurately. More qualitative results are presented in supplementary materials.  

\noindent \textbf{Comparison on Real-World Dataset.}
For the real-world dataset LFW-Ref-Test, the qualitative results in Figure~\ref{Fig:lfw} show that our framework can restore realistic and identity specific face images even if the reference ID images has different pose, expression and illumination with input LQ images. Whereas, PGDiff has the visual artifacts because of the gap between LQ and reference image, which may bring negative interference to the restoration. Meanwhile, FaceMe lags significantly behind in terms of image clarity.

\noindent \textbf{Robustness with different ID images.}
As shown in Figure \ref{fig:abl_ridfr_pgdiff}, given ID images with different poses, expressions, make-up, hair style and illumination, our framework is capable of generating high quality results with consistent identity and visual content, suppressing the interference from ID-irrelevant semantics, while PGDiff suffers from both low quality and inconsistency, and FaceMe suffers from instability with different ID images and lower ID fidelity.

\noindent \textbf{Robustness with low resolution ID images.}
To further demonstrate the robustness of our method, we down-sample all ID images to $128 \times 128$ or $64 \times 64$ to compare the performance on \textit{CelebRef-HQ-Test}. As shown in Table \ref{Tab:diff_quality} and Figure \ref{fig:different_resolution}, our method achieves consistent results among all three settings, showing robustness.

\begin{figure*}[ht]
    \centering
    \includegraphics[width=0.8\linewidth]{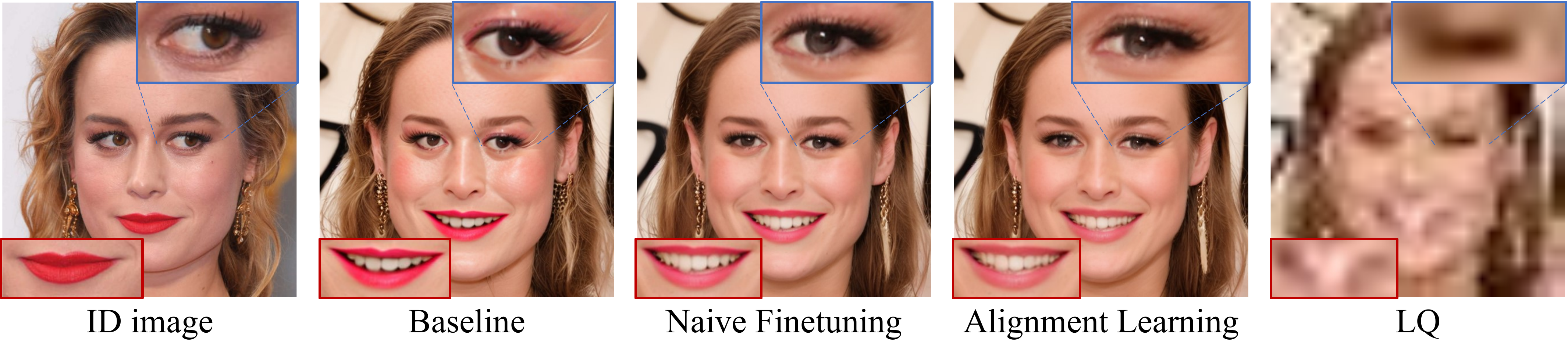}
    \vspace{-10pt}
    \caption{Ablation study on Alignment Learning, which has the most consistent result with LQ while maintaining identity fidelity.}
    \label{fig:abl_align_baseline}
    \vspace{-10pt}
\end{figure*}

\subsection{Ablation Study}
\noindent \textbf{Pixel-level Channel-wise Concatenation.}
\label{abl:concat}
We investigate the effectiveness of pixel-level channel-wise concatenation compared with latent-level channel-wise concatenation. We train two Content Injection Modules using different concatenation method mentioned in Section~\ref{cim} under the same training configuration. The results in Table~\ref{Tab:abl_concat} show the advantage in both LPIPS and image quality of pixel-level channel-wise concatenation.

\begin{table}[t]
    \caption{The effectiveness of pixel-level concatenation.}
    \label{Tab:abl_concat}
    \centering
    \resizebox{0.5\textwidth}{!}{
        \begin{tabular}{c|c c c c}
          \toprule
          Concatenation & LPIPS $\downarrow$ & MANIQA $\uparrow$ & MUSIQ $\uparrow$ \\
          \midrule
          Pixel-level (Ours) & \textbf{0.294} & \textbf{0.656} & \textbf{76.826} \\
          \midrule
          Latent-level & 0.417 &  0.651 & 76.384 \\
          \bottomrule
        \end{tabular}
    }
\end{table}

\noindent \textbf{Effectiveness of Alignment Learning.}
\label{abl:align}
To investigate the effectiveness of Alignment Learning and the robustness of our framework, we construct a synthetic dataset based on CelebRef-Ref-Test, examining results given the same $I_\text{LQ}$ with different $I_\text{ID}$ ($I_\text{LQ}$ and $I_\text{ID}$ share the same identity), both quantitatively and qualitatively.

For quantitative comparison, we propose a metric named Identity Variance (I.V.). Given the same $I_\text{LQ}$ , we calculate the variance of identity embeddings extracted from results under different reference images with the same identity. We compare three settings: baseline (no finetuning), naive finetuning using $<I_\text{LQ},I_\text{HQ},I_{\text{ID}}>$ pairs, and Alignment Learning. As shown in Table \ref{Tab:abl_align}, compared with baseline, naive finetuning can rougly align the results, while Alignment Learning leads to a more remarkable drop in variance.

\begin{table}[t]
    \caption{The effectiveness of Alignment Learning measured by Identity Variance (I.V.)}
    \label{Tab:abl_align}
    \centering
    \resizebox{0.5\textwidth}{!}{
        \begin{tabular}{c|c c c}
            \toprule
            & Baseline & Naive Finetuning  & Alignment Learning \\
            \midrule
            I.V. $\downarrow$ & 0.3610 & 0.2387 & 0.1447 \\
            \bottomrule
        \end{tabular}
    }
    \vspace{-2em}
\end{table}

For qualitative comparison, as shown in Figure \ref{fig:abl_align_baseline}, focusing on eye shape, eye direction and mouth, baseline is overly similar with ID image, indicating the interference of ID-irrelevant semantics. Naive finetuning improved overall result, while Alignment Learning achieves more consistent performance while maintaining identity fidelity.

\section{Conclusion}

\paragraph{Contribution} Diffusion priors are exploited as a prevalent approach in face restoration, in which the identity uncertainty issue has not been fully investigated and solved. In this study, we introduce RIDFR, a framework for ID-specific face restoration that focuses on leveraging pre-trained diffusion priors with both low-level content and high-level identity injection. Alignment Learning strategy is adopted to suppress the interference of ID-irrelevant face semantics. Comprehensive experiments demonstrate that our method excels in identity fidelity and visual quality with robust results.
\paragraph{Limitations}
Our framework only utilizes a single identity image during the inference stage, which ensures the practicality but makes it difficult to further refine the restoration results given multiple identity images. Utilizing more reference images has the potential to better suppress the interference of ID-irrelavant semantics. We will leave it to our future work.

\section{Acknowledgment}
This work is supported by National Natural Science Foundation of China (62271308, 62571322), STCSM (24ZR1432000, 24511106902, 24511106900, 22DZ2229005), 111 plan (BP0719010), and State Key Laboratory of UHD Video and Audio Production and Presentation.
%
%
%
\bibliographystyle{splncs04}
\bibliography{ref}
\end{document}